\def\year{2022}\relax
\documentclass[letterpaper]{article} % DO NOT CHANGE THIS
\usepackage{aaai22}  % DO NOT CHANGE THIS
\usepackage{times}  % DO NOT CHANGE THIS
\usepackage{helvet}  % DO NOT CHANGE THIS
\usepackage{courier}  % DO NOT CHANGE THIS
\usepackage[hyphens]{url}  % DO NOT CHANGE THIS
\usepackage{graphicx} % DO NOT CHANGE THIS
\usepackage{natbib}  % DO NOT CHANGE THIS AND DO NOT ADD ANY OPTIONS TO IT
\usepackage{caption} % DO NOT CHANGE THIS AND DO NOT ADD ANY OPTIONS TO IT
\DeclareCaptionStyle{ruled}{labelfont=normalfont,labelsep=colon,strut=off} % DO NOT CHANGE THIS
\usepackage{algorithm}
\usepackage{algorithmic}
\usepackage{epsfig}
\usepackage{verbatim}
\usepackage{tabularx}
\usepackage{array}
\usepackage{makecell}
\usepackage{placeins}
\usepackage{multirow, makecell}
\usepackage{amssymb}
\usepackage{amsmath}
\usepackage{newfloat}
\usepackage{listings}
\floatstyle{ruled}
\newfloat{listing}{tb}{lst}{}
\floatname{listing}{Listing}

\title{Integrated In-vehicle Monitoring System Using 3D Human Pose Estimation and Seat Belt Segmentation}
\author{
    Ginam Kim $^{\dagger}$, \textsuperscript{\rm 1} 
    Hyunsung Kim $^{\dagger}$, \textsuperscript{\rm 1} 
    Joseph Kihoon Kim, \textsuperscript{\rm 1} 
    Sung-Sik Cho, \textsuperscript{\rm 2}\\
    Yeong-Hun Park, \textsuperscript{\rm 2}
    Suk-Ju Kang, \textsuperscript{\rm 1}. \footnote{${\dagger}$ The first two authors are equally contributed for this paper.}
}
\affiliations{
    \textsuperscript{\rm 1} Sogang University\\
    \textsuperscript{\rm 2} Hyundai Mobis\\
    \{ginamkim, ghkskxlr, deepwork, sjkang\}@sogang.ac.kr\\
    \{sscho, yhpark0119\}@mobis.co.kr
}

\begin{document}

\maketitle

\begin{abstract}
Recently, along with interest in autonomous vehicles, the importance of monitoring systems for both drivers and passengers inside vehicles has been increasing. This paper proposes a novel in-vehicle monitoring system the combines 3D pose estimation, seat-belt segmentation, and seat-belt status classification networks. Our system outputs various information necessary for monitoring by accurately considering the data characteristics of the in-vehicle environment. Specifically, the proposed 3D pose estimation directly estimates the absolute coordinates of keypoints for a driver and passengers, and the proposed seat-belt segmentation is implemented by applying a structure based on the feature pyramid. In addition, we propose a classification task to distinguish between normal and abnormal states of wearing a seat belt using results that combine 3D pose estimation with seat-belt segmentation. These tasks can be learned simultaneously and operate in real-time. Our method was evaluated on a private dataset we newly created and annotated. The experimental results show that our method has significantly high performance that can be applied directly to real in-vehicle monitoring systems.
\end{abstract}

\section{Introduction}
Convolutional Neural Networks (CNNs) are widely applied to advanced driver assistance systems for autonomous driving \cite{chen2017multi, feng2020deep, liu2020importance, zhou2020joint}. These systems are generally used to process various information gathered from outside vehicles such as outside object detection and line segmentation. However, monitoring the conditions, behaviors, and seat-belt-wearing status of a driver and their passengers is very important to reduce the risk of accidents. In particular, the classification accuracy between normal and abnormal states of wearing a seat belt might help prevent fatalities or serious injury. However, existing in-vehicle monitoring systems have limitations in terms of classifying the condition, behavior, and seat-belt status of the driver and passengers. The CNNs in an in-vehicle monitoring system (IVMS) can simply solve these problems using a vision sensor. This paper proposes 3D human pose estimation to identify the conditions and behaviors of a driver and passengers and proposes a novel classification network for normal/abnormal seat-belt wearing. The results of our network can be adopted to give an alarm to passengers to improve safety.

\begin{figure}[t]
    \centering
    \includegraphics[width=\linewidth]{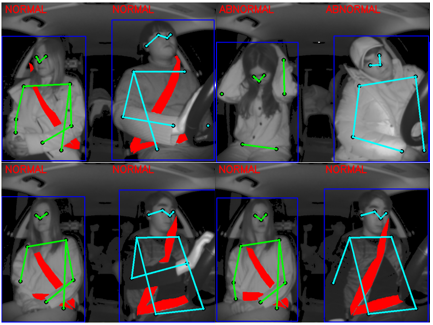}
    \caption{Overview of our dataset and proposed network.}
    \label{fig:fig1}
\end{figure}

Our architecture consists of the three following modules. First, we propose the absolute keypoints coordinate estimation method based on conventional 2D/3D human pose estimation networks \cite{xiao2018simple, moon2019camera}. Second, the proposed seat-belt segmentation network consists of parallel deconvolution structures. Third, the seat-belt wearing status classification is performed using the results of those two above mentioned networks and high-resolution features from the backbone network. The entire network is trained in an end-to-end manner, and it shows remarkable performance.

Generally, in-vehicle monitoring systems require an infrared (IR) camera to operate robustly regardless of the luminance change, unlike typical 3D human pose estimation. This means that a new dataset is necessary for in-vehicle monitoring because the IR dataset has different characteristics from typical RGB images. Additionally, since the 3D human pose dataset is generally produced under multi-view points, producing a new 3D human pose dataset has a high annotation cost. We solved this problem using the in-vehicle environment characteristics. Since previous datasets \cite{ionescu2013human3, mehta2018single} contain various positions of human objects, each image has a variety of root-depth. Therefore, rather than directly estimate the absolute depth of each keypoint, our method estimates the relative depth of each keypoint with an additional network that estimates the depth of the root keypoint. However, the variation of depth values in vehicles is limited. Furthermore, in most situations inside vehicles, this has almost a fixed value, unlike situations outside the vehicle. Therefore, each keypoint can be directly estimated without an additional root depth-estimating network. 

In these characteristics of in-vehicle monitoring, we annotate 2D keypoints using an infrared (IR) camera and depth values with a Time of Flight (ToF) camera. We use the depth value of the ToF camera as the ground truth depth. As a result, a 3D human pose dataset is produced with only a 2D keypoints annotation cost, thereby significantly reducing the annotation cost. Our private dataset includes consists of a total of 30,000 images. The contributions of this paper are summarized as follows.

\begin{itemize}
\item We propose a novel end-to-end network that integrates 3D human pose estimation, seat-belt segmentation, and seat-belt status classification. To our knowledge, 3D human pose estimation was first applied inside a vehicle.
\item A new insight for a data generation method is proposed to consider the characteristics of the vehicle's internal environment.
\item Our proposed method shows remarkable performance that can be directly applied to a real in-vehicle monitoring system that operates in real-time.
\end{itemize}

\section{Related Works}
\textbf{3D Human Pose Estimation}
3D human pose estimation is mainly categorized into top-down and bottom-up methods. Top-down methods use a cropped bounding box as input that contains a single person \cite{li20143d, sun2017compositional, pavlakos2017coarse, sun2018integral, moon2019camera, martinez2017simple, nie2017monocular, gong2021poseaug, llopart2020liftformer}. Meanwhile, bottom-up methods estimate all personal keypoints from the input image and then group them into each set of a person \cite{fabbri2020compressed, li2020hmor, lin2020hdnet, mehta2020xnect, wang2010combined}. Our proposed method taken the form of a top-down method. 

Top-down methods have two conventional approaches. One is the single-stage approach that directly estimates the 3D coordinates from an input cropped image \cite{li20143d, sun2017compositional, pavlakos2017coarse, sun2018integral, moon2019camera}. \cite{li20143d} trained regression and detection tasks simultaneously to obtain an accurate 3D human pose. \cite{sun2017compositional} adopted structure-aware regression, which showed that the regression-based method is more effective than the detection-based method for pose estimation. The network of \cite{pavlakos2017coarse} estimated the 3D human pose in a coarse-to-fine manner by applying CNNs iteratively. Therefore, the CNN refined the image features at every step. \cite{sun2018integral} proposed the soft-argmax operation to tackle issues caused by post-processing and quantization errors; this can be applied to any heatmap-based 3D pose estimation network to obtain coordinates with its differentiable property. \cite{moon2019camera} adopted the conventional idea of 2D multi-person top-down approaches to 3D multi-person pose estimation; they detected human bounding boxes using a human detector and then estimated the keypoints for each person. \cite{moon2019camera} used RootNet to estimate the absolute root location and PoseNet to estimate the root-relative coordinates; their method showed significant improvement in terms of 3D multi-person pose estimation.

The alternative is a two-stage approach with a lifting network \cite{martinez2017simple, nie2017monocular, llopart2020liftformer, gong2021poseaug}. The two-stage method first estimates 2D keypoints' coordinates and then translates 2D coordinates into 3D coordinates using an additional lifting network. \cite{martinez2017simple} proposed a simple and lightweight lifting network that could predict 3D human poses with given 2D keypoint locations. \cite{nie2017monocular} adopted Long Short-Term Memory (LSTM) to predict the depth of keypoints. The two types of LSTM used the results of 2D pose estimation and input image patches as input; they achieved better performance lifting 2D keypoints to 3D keypoints. \cite{llopart2020liftformer} used attention-based transformer encoder blocks to predict 3D keypoints; the inputs for this method were a sequence of 2D keypoints and the network generated 3D keypoints. \cite{gong2021poseaug} proposed an online augmentation method that could generate harder poses to estimate. Using the harder cases, the entire 3D pose estimation network learned various geometry factors of human poses.

Those two approaches generally estimate the depth value of the root keypoint and then the depth of each keypoint to add this to the root depth to produce the final output. Their adoption of this method lies in the characteristics of the dataset. The commonly used datasets \cite{ionescu2013human3, mehta2018single} have various depths of human objects in images. Some people exist nearby, and those who are far away also exist at the same time. Since the network cannot effectively estimate the wide depth range of the data, one keypoint (pelvis) is set as the root keypoint, the depth value of which is extracted by a separately designed network. Therefore, the keypoints estimation network estimates only the relative depth at each keypoint. This method showed effective performance.

\textbf{Human pose estimation for in-vehicle monitoring system} Recently developed 2D/3D human pose estimation networks using deep learning have shown remarkable performance. However, pose estimation networks for IVMS have not improved much. Only a few networks \cite{okuno2018body, yuen2018looking, chun2019nads, heo2020lightweight} have attempted to assess the performance in an in-vehicle environment, and even those have focused solely on 2D pose estimation. \cite{okuno2018body} proposed an architecture that estimated human pose and face orientation for an autonomous driving system that consisted of only three convolutional layers and a fully connected layer; through this shallow network, it can perform real-time processing. \cite{yuen2018looking} suggested predicting only the arms of the driver and passengers; this network used partial affinity fields (PAF) from \cite{cao2017realtime}. \cite{chun2019nads} has the most similar architecture to our proposed network; they performed 2D pose estimation and seat-belt segmentation and used PAF to estimate 2D keypoints, but they only estimated body keypoints without face keypoints.

\textbf{Seat belts} Efforts have been made to solve seat-belt-aware tasks such as detection, segmentation, and status classification in the area of computer vision, but trials to apply CNN remain in their infancy. \cite{zhou2017learning} tried to detect seat-belt by edge detection using a salient gradient. \cite{kashevnik2020seat} performed seat-belt status classification using Tiny-YOLO \cite{redmon2016you}. First, they detected the main part of the seat-belt and corner using Tiny-YOLO and then classified whether the seat-belt was fastened correctly. \cite{chun2019nads} (as mentioned above) performed seat-belt segmentation using a feature pyramid network during simultaneous 2D human pose estimation.

\section{Proposed Methods}
\begin{figure*}[t]
    \centering
    \includegraphics[width=\textwidth]{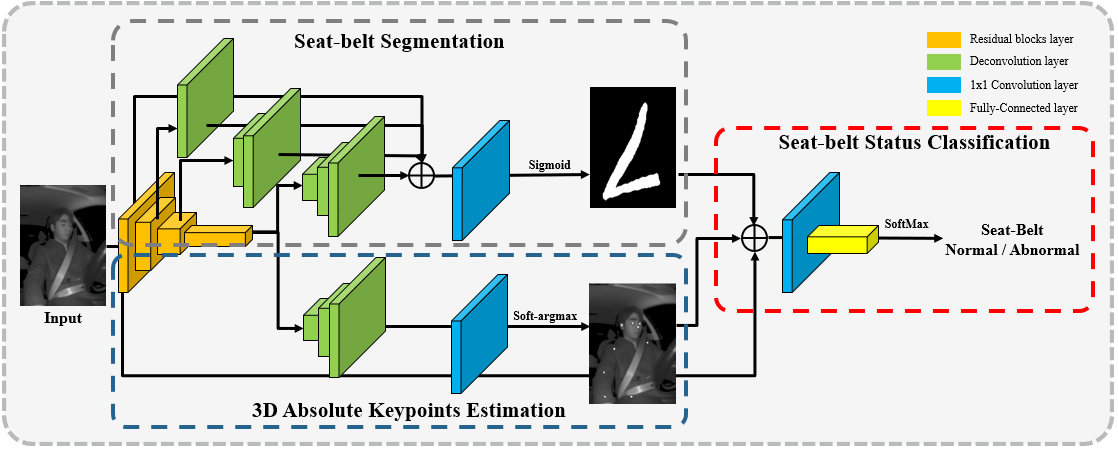}
    \caption{The overall architecture of the proposed system integrating 3D pose estimation, seat-belt segmentation, and seat-belt status classification.}
    \label{fig:architecture}
\end{figure*}

Our goal is to detect absolute 3D semantic keypoint coordinates of the driver and front passenger in a top-down manner and perform seat-belt segmentation using a single ToF camera. Finally, in this paper, our proposed network performs a seat-belt status classification. Figure \ref{fig:architecture} describes the overall architecture of the proposed method, which is composed of 3D pose estimation, seat-belt segmentation, and seat-belt classification. In the absolute 3D pose estimation, we extract the heatmaps of keypoints using the conventional CNN architecture. For the accurate seat-belt segmentation masks, we adopt the deconvolution layer-based parallel architecture to all output features in the backbone network and then use the output of those networks and the high-resolution feature as input. The following sections describe this in more detail.

\begin{table}[t]
\centering
\resizebox{\columnwidth}{!}{
\begin{tabular}{@{}c| c| c@{} }
\Xhline{3\arrayrulewidth}
    \multicolumn{3}{c}{Clothes}\\\hline
    Jacket, long-sleeve & short-sleeve & winter clothes\\\hline
    $34\%$ & $33\%$ & $33\%$\\\Xhline{3\arrayrulewidth}
\end{tabular}}
\caption{Subject statistics.}
\label{tab:statistics}
\end{table}

\subsection{Dataset generation}
The biggest bottleneck to applying CNN-based computer vision tasks in IVMS is appropriate training datasets. Few datasets are tailored to in-vehicle environments, so we manufactured a dataset to train our proposed network including 30K images. Moreover, we propose an efficient methodology to manufacture this dataset for the in-vehicle environment with relatively low cost. We set up IR and ToF cameras inside a vehicle to collect data on the driver and passengers. The ToF camera can collct and robustly operate depth information regardless of luminance changes. As summarized in Table \ref{tab:statistics}, the driver and passengers changed clothes several types to consider the situation of various seasons for almost 20 people. Each outfit accounts for 33\% of the total dataset. During data collection, we assumed various scenarios that may occur while driving. These scenarios include various general actions such as getting on and off, adjusting the seat position, operating an infotainment system, and operating a handle, as well as other actions such as stretching, and wearing a hat or a coat.

\begin{figure}[t]
    \centering
    \includegraphics[width=\linewidth]{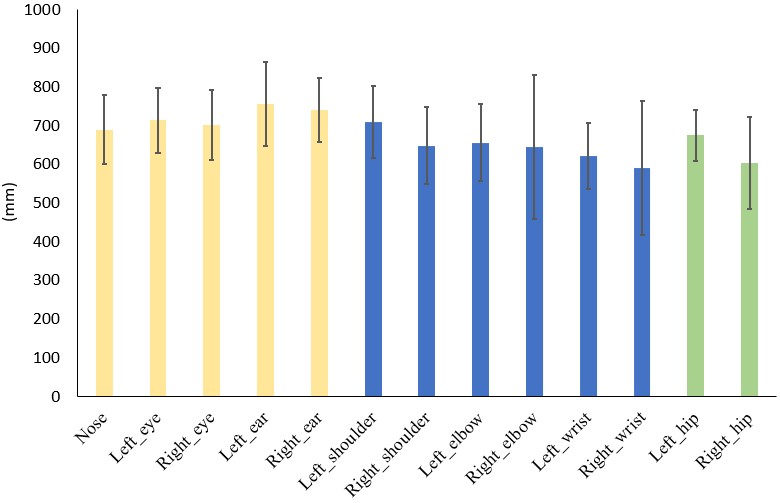}
    \caption{Depth value distribution of the keypoints}
    \label{fig:depth_distrib}
\end{figure}

\begin{figure}[t]
    \centering
    \includegraphics[width=\linewidth]{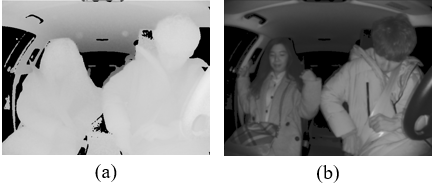}
    \caption{Collected image examples. (a) Normalized gray-scale image from the ToF camera. (b) Normalized gray-scale image from the IR camera.}
    \label{fig:normaliz}
\end{figure}

Our 3D absolute keypoints estimation network estimates the absolute 3D coordinates of keypoints from images cropped by detecting the human. In this case, the depth values for the driver and passengers in the vehicle are almost unchanged. Figure \ref{fig:depth_distrib} shows that most keypoints are distributed $400$--$900$ mm. In particular, $64.26\%$ of the keypoints exist within $500$--$800$ mm and $96.83\%$ are present within $400$--$900$mm. This means that the depth value variation is limited in the environment inside a vehicle and the process of estimating the root depth value using an additional root-depth estimation network is unnecessary. From this observation, we can predict the absolute coordinates without any additional root-depth estimation networks.

In addition, as shown in Figure \ref{fig:normaliz}(b), the image from the IR camera should be normalized for use as input. In the annotation process, we used the format of the MSCOCO dataset \cite{lin2014microsoft}, which is one of the most widely used datasets in object detection, semantic segmentation, and human pose estimation. Using this dataset, we first, made a bounding box for each person; thus, our dataset has only one object class (person). Second, 2D keypoint coordinates were annotated. The MSCOCO has 17 keypoints for every human, whereas, our dataset has only 13 points. 
In the in-vehicle environment, lower body parts are usually unseen; therefore we only collected the keypoints of the face and upper body. With the annotated 2D coordinates, we extracted the depth values at the same locations in the ToF raw data of Figure \ref{fig:normaliz}(a). Finally, we annotated the segmentation mask in the form of a polygon and divided the seat-belt status classes into normal or abnormal according to whether it is worn normally. A ToF camera was used for data generation. At the inference time, only the IR camera is used.

\subsection{Proposed Network Architecture}
\textbf{3D absolute keypoints estimation.} As described in Figure \ref{fig:architecture}, the 3D absolute keypoints estimation network is composed of a backbone network and three deconvolution layers. It is a simple architecture that is widely used in 2D/3D keypoints estimation \cite{xiao2018simple, moon2019camera}. We used ResNet50 \cite{he2016identity} as a backbone network. The extracted backbone feature $\textbf{F}_{B}$ becomes 3D keypoint heatmaps $\{{\mathbf H}_{k}\}_{k=1}^{K}$ after passing through the $4\times4$ deconvolution layer three times as follows:

\begin{equation} \label{1}
{\{{\mathbf H}_{k}\}_{k=1}^{K} = Deconv_{4\times4}^{3}(\textbf{F}_{B4})},
\end{equation}

\noindent where $K$ is the total number of keypoints. Since obtaining the coordinate of maximum value is a non-differentiable operation, the soft-argmax operation is used to obtain the 3D keypoint coordinates $\{[{x}, {y}, {z}]_{k}\}_{k=1}^{K}$ from $\{{\mathbf H}_{k}\}_{k=1}^{K}$ as follows:

\begin{equation} \label{2}
{\{[{x}, {y}, {z}]_{k}\}_{k=1}^{K} = Soft\_argmax(\{{\mathbf H}_{k}\}_{k=1}^{K})}.
\end{equation}

\textbf{Seat-belt segmentation.} Seat-belt segmentation predicts a binary segmentation mask. The binary mask from the ground truth polygons is used as a ground truth segmentation. To predict an accurate segmentation mask, the segmentation network has a parallel deconvolution layer structure that applies deconvolution to 2nd, 3rd, and 4th layer outputs of the backbone, respectively, and all features that are estimated as the backbone deepens can be used. Each deconvolution layer has the same kernel size of $4\times4$ and is applied differently depending on the resolution until reaching the same size as the output of the 1st layer. The upsampled features are concatenated with the output of the 1st layer and pass through the convolution layer once more and become ${F}_{seatbelt}$. Finally, the sigmoid function is used to extract the binary seat-belt segmentation mask ${Mask}_seatbelt$.

\begin{equation} \label{3}
{{Mask}_{seatbelt} = Sigmoid({F}_{seatbelt})},
\end{equation}

\textbf{Seat-belt status classification.} The seat-belt status classification network uses $\{{\mathbf H}_{k}\}_{k=1}^{K}$, ${F}_{seatbelt}$ and the high-resolution feature ${F}_{H}$ that comes from the first layer of the backbone as an input. Because $\{{\mathbf H}_{k}\}_{k=1}^{K}$ contains only heatmap information and  ${F}_{seatbelt}$ describes the seat-belt segmentation mask, ${F}_{H}$ is necessary to classify the seat-belt wearing status. Those features pass through the $1\times1$ convolution layer and a fully connected layer after being concatenated. Finally, with the softmax operation, the seat-belt status confidence score ${Cf}_{status}$ is generated.

\begin{table*}[t]
\centering
%\resizebox{\textwidth}{!}{
    \begin{tabular}{@{}c | c | c | c | c | c |c @{} }
    \Xhline{3\arrayrulewidth}
             & \multicolumn{2}{c|}{\multirow{2}{*}{Keypoints}} & \multirow{2}{*}{MPJPE(mm)} &  \multicolumn{3}{c}{Distribution(\%)}\\\cline{5-7}
            &           \multicolumn{2}{c|}{}                  & & $<$ 30mm & $<$ 50mm & $<$ 70mm \\\cline{1-7}
    \multirow{8}{*}{Driver} &\multirow{3}{*}{Face} & nose &  17.74 & 90.43 & 97.76 & 99.47 \\\cline{ 3-7}
                            & &eye      &  15.92 & 88.22 & 98.73 & 99.76 \\\cline{3-7}
                            & &ear      & 19.73 & 84.19 & 97.23 & 98.76 \\\cline{ 2-7}
                            &\multirow{3}{*}{Upper body}&shoulder & 33.96 & 53.93 & 85.35 & 96.02 \\\cline{ 3-7}
                            & & elbow      & 33.74 & 55.71 & 88.14 & 95.43 \\\cline{ 3-7}
                            & & wrist      & 56.83 & 50.29 & 71.88 & 80.46 \\\cline{ 2-7}
                            &Lower body& pelvis      & 53.82 & 36.08 & 70.97 & 88.68\\\cline{ 2-7}
                            & \multicolumn{2}{c|}{Total} & \multicolumn{4}{c}{31.14 mm}\\\Xhline{3\arrayrulewidth}

    \multirow{8}{*}{Passenger} &\multirow{3}{*}{Face} & nose & 33.95       & 58.76       & 79.55       & 91.64 \\\cline{ 3-7}
                            & &eye      & 56.13 & 41.13 & 64.05 & 77.93 \\\cline{3-7}
                            & &ear      & 60.01 & 22.80 & 49.76 & 74.47 \\\cline{ 2-7}
                            &\multirow{3}{*}{Upper body}&shoulder & 51.73 & 26.49 & 58.18 & 78.61 \\\cline{ 3-7}
                            & & elbow      & 54.72 & 31.09 & 60.16 & 77.67 \\\cline{ 3-7}
                            & & wrist      & 53.49 & 32.47 & 60.05 & 76.89 \\\cline{ 2-7}
                            &Lower body& pelvis      & 57.34 & 23.71 & 55.30 & 74.15\\\cline{ 2-7}
                            & \multicolumn{2}{c|}{Total} & \multicolumn{4}{c}{19.49 mm}\\\Xhline{3\arrayrulewidth}

    \multirow{8}{*}{Driver $\&$ Passenger} &\multirow{3}{*}{Face} & nose  & 25.16 & 75.26  & 88.29       & 95.89 \\\cline{ 3-7}
                            & &eye      & 33.57 & 67.55 & 83.51 & 90.17 \\\cline{3-7}
                            & &ear      & 24.88 & 76.34 & 91.16 & 95.82 \\\cline{ 2-7}
                            &\multirow{3}{*}{Upper body}&shoulder & 41.68 & 42.00 & 73.54 & 88.45 \\\cline{ 3-7}
                            & & elbow      & 46.40 & 40.86 & 71.26 & 84.71 \\\cline{ 3-7}
                            & & wrist      & 54.77 & 39.31 & 64.58 & 78.26 \\\cline{ 2-7}
                            &Lower body& pelvis      & 55.24 & 31.09 & 64.64 & 82.81\\\cline{ 2-7}
                            & \multicolumn{2}{c|}{Total} & \multicolumn{4}{c}{41.01 mm}\\\Xhline{3\arrayrulewidth}

    \Xhline{3\arrayrulewidth}
    \end{tabular}%}
    \caption{3D keypoints performance analysis on our dataset.}
    \label{tab:mpjpe}
\end{table*}

\subsection{Loss function}
We define the loss function for each task. The loss of 3D absolute keypoints estimation ${L}_{keypoints}$ is the Mean Absolute Error (MAE) which is calculated with $\{{\mathbf H}_{k}\}_{k=1}^{K}$ and the ground truth heatmap $\{{\mathbf Hgt}_{k}\}_{k=1}^{K}$. Moreover the seat-belt segmentation loss ${L}_{seg}$ and classification loss ${L}_{cls}$ are Mean Squared Error (MSE), respectively as follows:

\begin{equation} \label{4}
{{L}_{keypoints} = \frac{1}{n}\times\sum\left|\{{\mathbf H}_{k}\}_{k=1}^{K} - \{{\mathbf Hgt}_{k}\}_{k=1}^{K}\right|},
\end{equation}

\begin{equation} \label{5}
{{L}_{seg} = \frac{1}{n}\times{\sum\left|{Mask}_{seatbelt} - {Mask}_{gt}\right|}^2},
\end{equation}

\begin{equation} \label{6}
{{L}_{cls} = \frac{1}{n}\times{\sum\left|{Cf}_{status} - {Cf}_{gt}\right|}^2},
\end{equation}

\noindent where n is the total size of the data, ${Mask}_{gt}$ means the ground truth seat-belt segmentation mask and ${Cf}_{gt}$ is the ground truth one-hot vector of seat-belt status classes.
The total amount of loss is calculated as follows:

\begin{equation} \label{7}
{{L}_{total} = {L}_{keypoints} + \alpha{L}_{seg} + {L}_{cls}},
\end{equation}

\noindent where $\alpha$ is a hyper-parameter for ${L}_{seg}$. Using this loss function (\ref{7}), our entire proposed network can be trained in an end-to-end manner.

\section{Experiments}
\subsection{Implementation details}
 The proposed dataset contains $60,000$ person instances within $30,000$ images. We used 80\% of the generated dataset as a training set, and the other 20\% as a validation set. Our model was trained on the proposed training set without any extra data and experimental results were demonstrated on the validation set. The entire training and testing was performed with an NVIDIA GeForce RTX 3090 GPU. For the evaluation, the Mean Per Joint Position Error (MPJPE) is used as a 3D keypoints evaluation metric and the Interaction over Union (IoU) is employed as an evaluation metric for seat-belt segmentation. We used the Adam optimizer \cite{kingma2014adam} and the models were initialized randomly. In the training phase, the initial learning rate was set to $1$e\textminus$3$, and dropped to $1$e\textminus$4$ at the $50$th and $1$e\textminus$5$ at the $70$th epochs, respectively. ResNet50 \cite{he2016identity} was used as the backbone networks. We set $\alpha$ to 100.

\begin{figure*}[t]
    \centering
    \includegraphics[width=\textwidth]{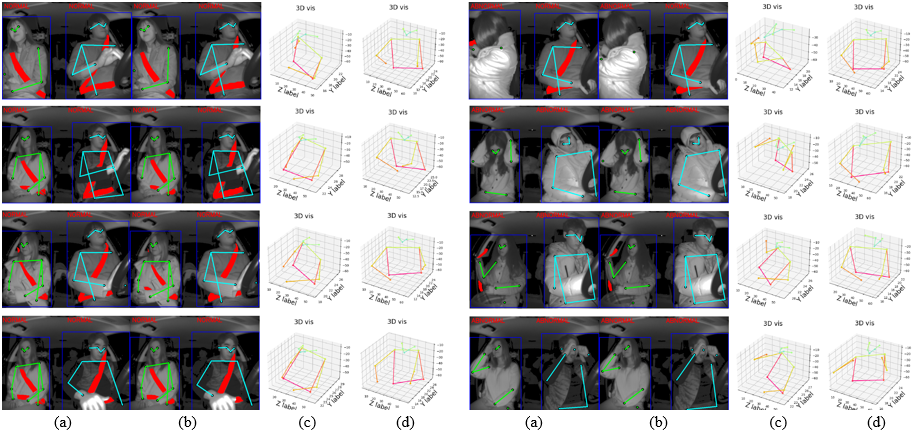}
    \caption{Estimated 3D human pose and seat-belt segmentation sample results. (a) ground truth, (b) estimated results, (c) 3D human pose estimation results of the passenger in the 3D domain, and (d) 3D human pose estimation results of the driver in 3D the domain.}
    \label{fig:qualitivity}
\end{figure*}

\begin{table}[]
\centering
\resizebox{\columnwidth}{!}{
\begin{tabular}{@{}c | c | c | c | c | c @{} }
\Xhline{3\arrayrulewidth}
    \multicolumn{6}{c}{Driver} \\\hline
    \multicolumn{3}{c|}{Left Keypoints MPJPE (mm)} & \multicolumn{3}{c}{Right Keypoints MPJPE (mm)}\\\hline
Face & Upper body & Pelvis & Face & Upper body & Pelvis \\\hline
18.49 &61.71 & 112.32 & 16.54 & 31.29 & 37.66\\\Xhline{3\arrayrulewidth}
    \multicolumn{6}{c}{Passenger} \\\hline
    \multicolumn{3}{c|}{Left Keypoints MPJPE (mm)} & \multicolumn{3}{c}{Right Keypoints MPJPE (mm)}\\\hline
Face & Upper body & Pelvis & Face & Upper body & Pelvis \\\hline
57.06 &47.68& 38.97 & 55.77 & 62.05 & 75.80\\\Xhline{3\arrayrulewidth}
\Xhline{3\arrayrulewidth}
\end{tabular}}
\caption{Comparison of the left and right body 3D keypoints MPJPE according to the driver and  the passenger.}
\label{tab:leftright}
\end{table}

\begin{table}[t]
\centering
\resizebox{\columnwidth}{!}{
\begin{tabular}{@{}c | c | c |c| c@{} }
\Xhline{3\arrayrulewidth}
    & 3D pose & Seat-belt & Seat-belt  & \multirow{2}{*}{Total}\\
    & estimation & segmentation & classification &\\\hline
    \multirow{2}{*}{Accuracy} & 41.01 mm & 80.64 \% & 95.90 \% & \multirow{2}{*}{-} \\
                              & (MPJPE) & (IoU) & (Accuracy) &\\\hline

    Speed (FPS) & 145.07 & 686.54 & 5824.67 & 129.03\\\Xhline{3\arrayrulewidth}
\end{tabular}}
\caption{Entire network performance evaluation.}
\label{tab:other}
\end{table}

\subsection{Results}
We analyzed the 3D pose estimation results as summarized in Table \ref{tab:mpjpe}; the results for the driver and front passenger were analyzed separately. When comparing the average values, the driver's MPJPE is 31.14mm, which is relatively lower than that of the passenger 52.26mm. Since we assumed actual driving situations when manufacturing the dataset, the driver concentrated on driving conditions and the passenger performed more malicious actions. The results for each keypoint show that overall, most keypoints were estimated to have an MPJPE within 70mm, and both the driver and passenger showed a lower MPJPE for the face keypoints than the upper body keypoints. In Table \ref{tab:leftright}, a remarkable point is that the driver has a higher error in the left keypoints of their body than in the right, while the passenger shows the opposite. From these results, we can analyze that estimating the outside keypoints of both people is more complicated because outside keypoints are more vulnerable to occlusion due to the camera's angle of view and several objects. The MPJPE for the entire test set is 41.01 mm; it shows better performance than state-of-the-art networks have achieved in public datasets. These results prove that our proposed network is sufficiently effective to be applied directly in in-vehicle environments.

As summarized in Table \ref{tab:other}, we evaluated the overall network performance. As mentioned above, the 3D pose estimation performance shows an MPJPE of 41.01 mm, and the 3D pose estimation network operates at 145.07 fps. Seat-belt segmentation also has a high IoU performance of 80.64\% and 686.54 fps in a single operation. Finally, the seat-belt classification shows high accuracy of 95.90\%. The operation speed of the entire network is 129.03 fps using an NVIDIA 3090 RTX. As described in Figure \ref{fig:qualitivity}, the qualitative results of our proposed network show remarkable performance. Our method implements seat-belt segmentation precisely even when little of the seat-belt is visible. The human pose reconstructed in 3D implies that our method could be applied to detect abnormal postures in vehicles. This proves that our proposed network is efficient at constructing a 3D human pose in in-vehicle conditions.

\section{Conclusion}
We proposed a novel method for an in-vehicle monitoring system for drivers and passengers. We first suggested an efficient methodology to manufacture an in-vehicle-aware dataset. Many conditions of in-vehicle environments were limited in terms of the area, number, and size of human objects and the movement of humans. Therefore producing datasets that consider these limitations can lower the annotation cost. We demonstrated the effectiveness of our method by applying it to our proposed network, which is a novel integrated framework that uses the 3D human pose estimation, seat-belt segmentation, and seat-belt status classification. Moreover, those tasks can be trained in an end-to-end manner.
We believe that this study provides a novel solution for the in-vehicle monitoring of advanced driver assistance systems and thus enhances the safety for humans.
%\clearpage
\bibliography{aaai22}

\begin{thebibliography}{32}
\providecommand{\natexlab}[1]{#1}

\bibitem[{Cao et~al.(2017)Cao, Simon, Wei, and Sheikh}]{cao2017realtime}
Cao, Z.; Simon, T.; Wei, S.-E.; and Sheikh, Y. 2017.
\newblock Realtime multi-person 2d pose estimation using part affinity fields.
\newblock In \emph{Proceedings of the IEEE conference on computer vision and
  pattern recognition (CVPR)}, 7291--7299.

\bibitem[{Chen et~al.(2017)Chen, Ma, Wan, Li, and Xia}]{chen2017multi}
Chen, X.; Ma, H.; Wan, J.; Li, B.; and Xia, T. 2017.
\newblock Multi-view 3d object detection network for autonomous driving.
\newblock In \emph{Proceedings of the IEEE conference on Computer Vision and
  Pattern Recognition (CVPR)}, 1907--1915.

\bibitem[{Chun et~al.(2019)Chun, Hamidi~Ghalehjegh, Choi, Schwarz, Gaspar,
  McGehee, and Baek}]{chun2019nads}
Chun, S.; Hamidi~Ghalehjegh, N.; Choi, J.; Schwarz, C.; Gaspar, J.; McGehee,
  D.; and Baek, S. 2019.
\newblock NADS-Net: A Nimble Architecture for Driver and Seat Belt Detection
  via Convolutional Neural Networks.
\newblock In \emph{IEEE/CVF International Conference on Computer Vision
  Workshops (ICCVW)}, 0--0.

\bibitem[{Fabbri et~al.(2020)Fabbri, Lanzi, Calderara, Alletto, and
  Cucchiara}]{fabbri2020compressed}
Fabbri, M.; Lanzi, F.; Calderara, S.; Alletto, S.; and Cucchiara, R. 2020.
\newblock Compressed volumetric heatmaps for multi-person 3d pose estimation.
\newblock In \emph{Proceedings of the IEEE/CVF Conference on Computer Vision
  and Pattern Recognition (CVPR)}, 7204--7213.

\bibitem[{Feng et~al.(2020)Feng, Haase-Sch{\"u}tz, Rosenbaum, Hertlein,
  Glaeser, Timm, Wiesbeck, and Dietmayer}]{feng2020deep}
Feng, D.; Haase-Sch{\"u}tz, C.; Rosenbaum, L.; Hertlein, H.; Glaeser, C.; Timm,
  F.; Wiesbeck, W.; and Dietmayer, K. 2020.
\newblock Deep multi-modal object detection and semantic segmentation for
  autonomous driving: Datasets, methods, and challenges.
\newblock \emph{IEEE Transactions on Intelligent Transportation Systems},
  22(3): 1341--1360.

\bibitem[{Gong, Zhang, and Feng(2021)}]{gong2021poseaug}
Gong, K.; Zhang, J.; and Feng, J. 2021.
\newblock PoseAug: A Differentiable Pose Augmentation Framework for 3D Human
  Pose Estimation.
\newblock In \emph{Proceedings of the IEEE/CVF Conference on Computer Vision
  and Pattern Recognition (CVPR)}, 8575--8584.

\bibitem[{He et~al.(2016)He, Zhang, Ren, and Sun}]{he2016identity}
He, K.; Zhang, X.; Ren, S.; and Sun, J. 2016.
\newblock Identity mappings in deep residual networks.
\newblock In \emph{Proceedings of the European conference on computer vision
  (ECCV)}, 630--645.

\bibitem[{Heo et~al.(2020)Heo, Kim, Park, Kim, Cho, Lee, and
  Kang}]{heo2020lightweight}
Heo, J.; Kim, G.; Park, J.; Kim, Y.; Cho, S.-S.; Lee, C.~W.; and Kang, S.-J.
  2020.
\newblock Lightweight Deep Neural Network-based Real-Time Pose Estimation on
  Embedded Systems.
\newblock In \emph{Proceedings of the IEEE Intelligent Vehicles Symposium
  (IV)}, 1066--1071.

\bibitem[{Ionescu et~al.(2013)Ionescu, Papava, Olaru, and
  Sminchisescu}]{ionescu2013human3}
Ionescu, C.; Papava, D.; Olaru, V.; and Sminchisescu, C. 2013.
\newblock Human3. 6m: Large scale datasets and predictive methods for 3d human
  sensing in natural environments.
\newblock \emph{IEEE transactions on pattern analysis and machine intelligence
  (TPAMI)}, 36(7): 1325--1339.

\bibitem[{Kashevnik et~al.(2020)Kashevnik, Ali, Lashkov, and
  Shilov}]{kashevnik2020seat}
Kashevnik, A.; Ali, A.; Lashkov, I.; and Shilov, N. 2020.
\newblock Seat Belt Fastness Detection Based on Image Analysis from Vehicle
  In-abin Camera.
\newblock In \emph{Conference of Open Innovations Association (FRUCT)},
  143--150.

\bibitem[{Kingma and Ba(2014)}]{kingma2014adam}
Kingma, D.~P.; and Ba, J. 2014.
\newblock Adam: A method for stochastic optimization.
\newblock \emph{arXiv preprint arXiv:1412.6980}.

\bibitem[{Li et~al.(2020)Li, Wang, Liu, Qian, and Lu}]{li2020hmor}
Li, J.; Wang, C.; Liu, W.; Qian, C.; and Lu, C. 2020.
\newblock Hmor: Hierarchical multi-person ordinal relations for monocular
  multi-person 3d pose estimation.
\newblock \emph{arXiv preprint arXiv:2008.00206}.

\bibitem[{Li and Chan(2014)}]{li20143d}
Li, S.; and Chan, A.~B. 2014.
\newblock 3d human pose estimation from monocular images with deep
  convolutional neural network.
\newblock In \emph{Proceedings of the Asian Conference on Computer Vision
  (ACCV)}, 332--347.

\bibitem[{Lin and Lee(2020)}]{lin2020hdnet}
Lin, J.; and Lee, G.~H. 2020.
\newblock Hdnet: Human depth estimation for multi-person camera-space
  localization.
\newblock In \emph{Proceedings of the European Conference on Computer Vision
  (ECCV)}, 633--648.

\bibitem[{Lin et~al.(2014)Lin, Maire, Belongie, Hays, Perona, Ramanan,
  Doll{\'a}r, and Zitnick}]{lin2014microsoft}
Lin, T.-Y.; Maire, M.; Belongie, S.; Hays, J.; Perona, P.; Ramanan, D.;
  Doll{\'a}r, P.; and Zitnick, C.~L. 2014.
\newblock Microsoft coco: Common objects in context.
\newblock In \emph{Proceedings of the European conference on computer vision
  (ECCV)}, 740--755.

\bibitem[{Liu et~al.(2020)Liu, Han, Bai, Ge, Wang, Han, Li, You, and
  Lu}]{liu2020importance}
Liu, X.; Han, Y.; Bai, S.; Ge, Y.; Wang, T.; Han, X.; Li, S.; You, J.; and Lu,
  J. 2020.
\newblock Importance-aware semantic segmentation in self-driving with discrete
  wasserstein training.
\newblock In \emph{Proceedings of the AAAI Conference on Artificial
  Intelligence}, volume~34, 11629--11636.

\bibitem[{Llopart(2020)}]{llopart2020liftformer}
Llopart, A. 2020.
\newblock LiftFormer: 3D Human Pose Estimation using attention models.
\newblock \emph{arXiv preprint arXiv:2009.00348}.

\bibitem[{Martinez et~al.(2017)Martinez, Hossain, Romero, and
  Little}]{martinez2017simple}
Martinez, J.; Hossain, R.; Romero, J.; and Little, J.~J. 2017.
\newblock A simple yet effective baseline for 3d human pose estimation.
\newblock In \emph{Proceedings of the IEEE International Conference on Computer
  Vision (CVPR)}, 2640--2649.

\bibitem[{Mehta et~al.(2020)Mehta, Sotnychenko, Mueller, Xu, Elgharib, Fua,
  Seidel, Rhodin, Pons-Moll, and Theobalt}]{mehta2020xnect}
Mehta, D.; Sotnychenko, O.; Mueller, F.; Xu, W.; Elgharib, M.; Fua, P.; Seidel,
  H.-P.; Rhodin, H.; Pons-Moll, G.; and Theobalt, C. 2020.
\newblock XNect: Real-time multi-person 3D motion capture with a single RGB
  camera.
\newblock \emph{ACM Transactions on Graphics (TOG)}, 39(4): 82--1.

\bibitem[{Mehta et~al.(2018)Mehta, Sotnychenko, Mueller, Xu, Sridhar,
  Pons-Moll, and Theobalt}]{mehta2018single}
Mehta, D.; Sotnychenko, O.; Mueller, F.; Xu, W.; Sridhar, S.; Pons-Moll, G.;
  and Theobalt, C. 2018.
\newblock Single-shot multi-person 3d pose estimation from monocular rgb.
\newblock In \emph{International Conference on 3D Vision (3DV)}, 120--130.

\bibitem[{Moon, Chang, and Lee(2019)}]{moon2019camera}
Moon, G.; Chang, J.~Y.; and Lee, K.~M. 2019.
\newblock Camera distance-aware top-down approach for 3d multi-person pose
  estimation from a single rgb image.
\newblock In \emph{Proceedings of the IEEE/CVF International Conference on
  Computer Vision (ICCV)}, 10133--10142.

\bibitem[{Nie, Wei, and Zhu(2017)}]{nie2017monocular}
Nie, B.~X.; Wei, P.; and Zhu, S.-C. 2017.
\newblock Monocular 3d human pose estimation by predicting depth on joints.
\newblock In \emph{Proceedings of the IEEE International Conference on Computer
  Vision (ICCV)}, 3467--3475.

\bibitem[{Okuno et~al.(2018)Okuno, Yamashita, Fukui, Noridomi, Arata, Yamauchi,
  and Fujiyoshi}]{okuno2018body}
Okuno, K.; Yamashita, T.; Fukui, H.; Noridomi, S.; Arata, K.; Yamauchi, Y.; and
  Fujiyoshi, H. 2018.
\newblock Body posture and face orientation estimation by convolutional network
  with heterogeneous learning.
\newblock In \emph{International Workshop on Advanced Image Technology
  (IWAIT)}, 1--4.

\bibitem[{Pavlakos et~al.(2017)Pavlakos, Zhou, Derpanis, and
  Daniilidis}]{pavlakos2017coarse}
Pavlakos, G.; Zhou, X.; Derpanis, K.~G.; and Daniilidis, K. 2017.
\newblock Coarse-to-fine volumetric prediction for single-image 3D human pose.
\newblock In \emph{Proceedings of the IEEE conference on computer vision and
  pattern recognition (CVPR)}, 7025--7034.

\bibitem[{Redmon et~al.(2016)Redmon, Divvala, Girshick, and
  Farhadi}]{redmon2016you}
Redmon, J.; Divvala, S.; Girshick, R.; and Farhadi, A. 2016.
\newblock You only look once: Unified, real-time object detection.
\newblock In \emph{Proceedings of the IEEE conference on computer vision and
  pattern recognition (CVPR)}, 779--788.

\bibitem[{Sun et~al.(2017)Sun, Shang, Liang, and Wei}]{sun2017compositional}
Sun, X.; Shang, J.; Liang, S.; and Wei, Y. 2017.
\newblock Compositional human pose regression.
\newblock In \emph{Proceedings of the IEEE International Conference on Computer
  Vision (ECCV)}, 2602--2611.

\bibitem[{Sun et~al.(2018)Sun, Xiao, Wei, Liang, and Wei}]{sun2018integral}
Sun, X.; Xiao, B.; Wei, F.; Liang, S.; and Wei, Y. 2018.
\newblock Integral human pose regression.
\newblock In \emph{Proceedings of the European Conference on Computer Vision
  (ECCV)}, 529--545.

\bibitem[{Wang et~al.(2010)Wang, Ai, Yamashita, and Lao}]{wang2010combined}
Wang, S.; Ai, H.; Yamashita, T.; and Lao, S. 2010.
\newblock Combined top-down/bottom-up human articulated pose estimation using
  Adaboost learning.
\newblock In \emph{Proceedings of the International Conference on Pattern
  Recognition (ICPR)}, 3670--3673.

\bibitem[{Xiao, Wu, and Wei(2018)}]{xiao2018simple}
Xiao, B.; Wu, H.; and Wei, Y. 2018.
\newblock Simple baselines for human pose estimation and tracking.
\newblock In \emph{Proceedings of the European conference on computer vision
  (ECCV)}, 466--481.

\bibitem[{Yuen and Trivedi(2018)}]{yuen2018looking}
Yuen, K.; and Trivedi, M.~M. 2018.
\newblock Looking at hands in autonomous vehicles: A convnet approach using
  part affinity fields.
\newblock \emph{arXiv preprint arXiv:1804.01176}.

\bibitem[{Zhou et~al.(2017)Zhou, Chen, Tian, and Peng}]{zhou2017learning}
Zhou, B.; Chen, L.; Tian, J.; and Peng, Z. 2017.
\newblock Learning-based seat belt detection in image using salient gradient.
\newblock In \emph{Proceedings of the IEEE Conference on Industrial Electronics
  and Applications (ICIEA)}, 547--550.

\bibitem[{Zhou et~al.(2020)Zhou, Fang, Song, Liu, Yin, Dai, Li, and
  Yang}]{zhou2020joint}
Zhou, D.; Fang, J.; Song, X.; Liu, L.; Yin, J.; Dai, Y.; Li, H.; and Yang, R.
  2020.
\newblock Joint 3d instance segmentation and object detection for autonomous
  driving.
\newblock In \emph{Proceedings of the IEEE/CVF Conference on Computer Vision
  and Pattern Recognition}, 1839--1849.

\end{thebibliography}

\end{document}